# Vulnerability of deep neural networks for detecting COVID-19 cases from chest X-ray images to universal adversarial attacks

**Hokuto Hirano[1], Kazuki Koga[1], Kazuhiro Takemoto[1*]**
*1) Department of Bioscience and Bioinformatics, Kyushu Institute of Technology, Iizuka, Fukuoka 820-8502, Japan*
*\*Corresponding author's e-mail: takemoto@bio.kyutech.ac.jp*

## Abstract

Under the epidemic of the novel coronavirus disease 2019 (COVID-19), chest X-ray computed tomography imaging is being used for effectively screening COVID-19 patients. The development of computer-aided systems based on deep neural networks (DNNs) has been advanced, to rapidly and accurately detect COVID-19 cases, because the need for expert radiologists, who are limited in number, forms a bottleneck for the screening. However, so far, the vulnerability of DNN-based systems has been poorly evaluated, although DNNs are vulnerable to a single perturbation, called universal adversarial perturbation (UAP), which can induce DNN failure in most classification tasks. Thus, we focus on representative DNN models for detecting COVID-19 cases from chest X-ray images and evaluate their vulnerability to UAPs generated using simple iterative algorithms. We consider nontargeted UAPs, which cause a task failure resulting in an input being assigned an incorrect label, and targeted UAPs, which cause the DNN to classify an input into a specific class. The results demonstrate that the models are vulnerable to nontargeted and targeted UAPs, even in case of small UAPs. In particular, 2% norm of the UPAs to the average norm of an image in the image dataset achieves >85% and >90% success rates for the nontargeted and targeted attacks, respectively. Due to the nontargeted UAPs, the DNN models judge most chest X-ray images as COVID-19 cases. The targeted UAPs make the DNN models classify most chest X-ray images into a given target class. The results indicate that careful consideration is required in practical applications of DNNs to COVID-19 diagnosis; in particular, they emphasize the need for strategies to address security concerns. As an example, we show that iterative fine-tuning of the DNN models using UAPs improves the robustness of the DNN models against UAPs.

# 1. Introduction

Coronavirus disease 2019 (COVID-19) [1] is an infectious disease caused by the coronavirus, called severe acute respiratory syndrome coronavirus 2. The epidemic of COVID-19 started from Wuhan, China [2], and has had a serious impact on public health and economy globally [3]. To reduce the spread of this epidemic, effective screening of COVID-19 patients is required. For this purpose, positive real-time polymerase chain reaction (PCR) tests are mainly used [4]; however, they are often time-consuming, laborious, and involve complicated manual processes. Chest radiography imaging, especially chest X-ray computed tomography (CT) imaging, is an alternative screening method [5], because patients present abnormalities in chest radiography images, which are characteristic of those infected with COVID-19 [2,6]. Moreover, there are advantages to leveraging chest X-ray imaging for COVID-19 screening amid the pandemic in terms of rapid triaging, portability, availability, and accessibility [7]. However, the visual differences in chest X-ray images among COVID-19-associated pneumonia, non-COVID-19 pneumonia, and no pneumonia are subtle; thus, the need for expert radiologists, who are limited in number, forms a bottleneck for diagnoses based on radiography images. To avoid this limitation, computer-aided systems that can aid radiologists to more rapidly and accurately interpret radiography images to detect COVID-19 cases are highly desired [7,8]; in particular, deep neural networks (DNNs) are often used for this purpose.

DNNs are widely used for image classification, a task in which an input image is assigned a class from a fixed set of classes, as well as medical science [9,10], including diagnoses based on radiography images. Specifically, DNN-based systems can detect subtle visual differences in the images; in particular, a DNN can accurately distinguish bacterial and viral pneumonia in chest X-ray images [11]. Inspired by these previous studies, many researchers have constructed large-scale datasets of chest radiography images on COVID-19 [7,8,12,13] and have proposed DNN-based systems for screening COVID-19 cases from these images [8,14–17]. However, DNN-based systems in medical science have generally been closed-source and unavailable to the research community for deeper understanding and extension. Thus, Wang et al. [7] proposed COVID-Net, a deep convolutional neural network design tailored to detect COVID-19 cases from chest X-ray images, which is open-source and available to the general public. As the authors mentioned [7], this study will be leveraged and build upon by both researchers and citizen data scientists to accelerate the development of highly accurate yet practical deep learning solutions for detecting COVID-19 cases and accelerate COVID-19 treatment. The COVID-Net models are intended to be used as reference models; in fact, several DNN-based systems [18–20] for detecting COVID-19 cases have already been proposed, inspired by the COVID-Net study.

However, previous studies have poorly evaluated the vulnerabilities in DNNs, although DNNs are known to be vulnerable to adversarial examples [21,22], which are input images that cause misclassifications by DNNs and usually generated by adding specific, imperceptible perturbations to original input images that have been correctly classified using DNNs. Notably, a single perturbation (called *universal adversarial perturbation; UAP*) [23] that can induce DNN failure in most image classification tasks is also known to exist. Moreover, DNNs are less robust to not only nontargeted attacks [23], which cause misclassification (i.e., a task failure resulting in an input image being assigned an incorrect class), but also targeted attacks [24], which cause the DNN to classify an input image into a specific class. The existence of adversarial examples questions the generalization ability of DNNs, reduces model interpretability, and limits the applications of deep learning in safety- and security-critical environments [25]. Specifically, vulnerability is a serious problem in medical diagnosis [26]. Thus, it is important to evaluate how vulnerable the proposed DNN-based systems are to adversarial attacks (attacks based on UAPs, in particular) in practical applications. In addition, defense strategies against adversarial attacks (i.e., adversarial defense [22]) are required.

In this study, we focus on the COVID-Net models, which are representative models for detecting COVID-19 cases from chest X-ray images, and aim to evaluate the vulnerability of DNNs to adversarial attacks; specifically, the vulnerability to nontargeted and targeted attacks, based on UAPs, is investigated. Moreover, adversarial defense is considered; in particular, we evaluate how much the robustness of COVID-Net models to nontargeted and targeted UAPs increases using adversarial retraining [23,27] (i.e., fine-tuning with adversarial images).

## 2. Material and methods

### 2.1. COVID-Net models

We forked the COVID-Net repository (github.com/lindawangg/COVID-Net) on May 1, 2020, and obtained two DNN models for detecting COVID-19 cases from chest X-ray images: COVIDNet-CXR Small and COVIDNet-CXR Large. Moreover, we downloaded the COVIDx dataset, a collection of chest radiography images from several open-source chest radiography datasets, on May 1, 2020, according to the description in the COVID-Net repository. The chest X-ray images in the dataset were classified into three classes: *normal* (no pneumonia), *pneumonia* (non-COVID19 pneumonia; e.g., viral and bacterial pneumonia), and *COVID-19* (COVID-19 viral pneumonia). The dataset comprised 13,569 training images (7,966 *normal* images, 5,451 *pneumonia* images, and 152 *COVID-19* images) and 231 test images (100 *normal* images, 100 *pneumonia* images, and 31 *COVID-19* images).

### 2.2. Universal adversarial perturbations

The UAPs for nontargeted and targeted attacks were generated using a simple iterative algorithm [23,24]. The algorithm considers a classifier, $C(x)$, which returns the class or label with the highest confidence score for an input image, $x$. The algorithm starts with $\rho = 0$ (no perturbation) and iteratively updates the UAP, $\rho$, under the constraint that the $L_p$ norm of the perturbation is equal to or less than a small $\xi$ value (i.e., $\|\rho\|_p \leq \xi$), by additively obtaining an adversarial perturbation for an input image, $x$, which is randomly selected from an input image set, $X$, without replacement. These iterative updates continue till the number of iterations reaches the maximum $i_{\max}$.

The fast gradient sign method (FGSM) [21] was used to obtain an adversarial perturbation for the input image, whereas the original UAP algorithm [23] uses the DeepFool method [28]. This is because FGSM is used for both nontargeted and targeted attacks, and DeepFool requires a higher computational cost than FGSM and only generates a nontargeted adversarial example for the input image. FGSM generates the adversarial perturbation, $\hat{\rho}$, for $x$ using gradient $\nabla_x L(x, y)$ of the loss function at the given image $x$ and class $y$ with respect to the pixels [21]. For the $L_\infty$ norm, a nontargeted perturbation that causes misclassification is computed as $\hat{\rho} = \epsilon \cdot \text{sign}(\nabla_x L(x, C(x)))$, whereas a targeted perturbation that causes $C$ to classify image $x$ into class $y$ is obtained as $\hat{\rho} = -\epsilon \cdot \text{sign}(\nabla_x L(x, y))$, where $\epsilon$ ($> 0$) is the attack strength. For the $L_1$ and $L_2$ norms, a nontargeted perturbation is computed as $\hat{\rho} = \epsilon \cdot \nabla_x L(x, C(x)) / \|\nabla_x L(x, C(x))\|_p$, whereas a targeted perturbation is obtained as $\hat{\rho} = -\epsilon \cdot \nabla_x L(x, y) / \|\nabla_x L(x, y)\|_p$.

In the algorithms, FGSM is performed based on output $C(x + \rho)$ of the classifier for the perturbated image, $x + \rho$, at each iteration step. For nontargeted (targeted) attacks, an adversarial perturbation, $\hat{\rho}$, for $x + \rho$ is obtained using FGMS if $C(x + \rho) = C(x)$ ($C(x + \rho) \neq y$). After generating the adversarial example (i.e., $x_{\text{adv}} \leftarrow x + \rho + \hat{\rho}$) at this step, perturbation $\rho$ is updated if $C(x_{\text{adv}}) \neq C(x)$ ($C(x_{\text{adv}}) = y$) for nontargeted (targeted) attacks. When updating $\rho$, a projection function project, $(x, p, \xi)$, is used to satisfy the constraint that $\|\rho\|_p \leq \xi$ : $\rho \leftarrow \text{project}(x_{\text{adv}} - x, p, \xi)$, where $\text{poject}(x, p, \xi) = \arg\min_{x'} \|x - x'\|_2$ subject to $\|\rho\|_p \leq \xi$.

The details of the algorithms are described in [23,24]. We used the nontargeted UAP algorithm available in the Adversarial Robustness 360 Toolbox (ART) [29] (version 1.0; github.com/IBM/adversarial-robustness-toolbox). The targeted UAP algorithm was implemented by modifying the nontargeted UAP algorithm in ART in our previous study [24] (github.com/hkthirano/targeted_UAP_CIFAR10).

The nontargeted and targeted UAPs were generated using 13,569 training images in the COVIDx dataset. Parameter $\epsilon$ was set to 0.001; the cases where $p = 2$ and $\infty$ were considered. Parameter $\xi$ was determined based on ratio $\zeta$ of the $L_p$ norm of the UAP to the average $L_p$ norm of an image in the COVIDx dataset. The cases in which $\zeta = 1\%$ and 2% (i.e., almost imperceptible perpetuations) were considered. The average $L_\infty$ and $L_2$ norms were 237 and 32,589, respectively; $i_{\max}$ was set to 15.

For comparing the performance of the generated UAPs with that of random controls, we also generated random vectors (random UAPs) sampled uniformly from the sphere of a given radius [23].

## 2.3. Vulnerability evaluation

To evaluate the vulnerability of the DNN models to UAPs, we used the fooling rate, $R_f$, and targeted the attack success rate, $R_s$, of nontargeted and targeted attacks, respectively. $R_f$ for an image set is defined as the proportion of images that were not classified into their associated actual labels to all images in the set. $R_s$ for an image set is the proportion of adversarial images classified into the target class to all images in the set. Additionally, we obtained the confusion matrices to evaluate the change in prediction due to UAPs for each class (infection type).

*2.4. Adversarial retraining*

We performed adversarial retraining to increase the robustness of the COVID-Net models to UAPs [23,27]; in particular, the models were fine-tuned with adversarial images. The procedure was described in a previous study [23]. A brief description is as follows. 1) Ten UAPs against a DNN model were generated using the algorithm (for generating a nontargeted or targeted UAP) (Sec. 2.2) with the (clean) training image set. 2) A modified training image set was obtained by randomly selecting half of the training images and combining them with the rest in which each image was perturbed by a UAP randomly selected from 10 UAPs. 3) The model was fine-tuned by performing 5 extra epochs of training on the modified training image set. 4) A new UAP (against the fine-tuned model) was generated using the algorithm with the training image set. 5) Then, $R_f$ and $R_s$ of the UAP for the test images were computed. Steps 1)–5) were repeated five times.

## 3. Results

*3.1. Performance of COVID-Net models*

The test accuracies of the COVIDNet-CXR Small and COVIDNet-CXR Large models were 92.6% and 94.4%, respectively, and their training accuracies were 95.8% and 94.1%, respectively. As shown in the COVID-Net study [7], we also confirmed that the COVID-Net models achieved good accuracies.

*3.2. Vulnerability to nontargeted universal adversarial perturbations*

However, we found that both COVIDNet-CXR Small and COVIDNet-CXR Large models were vulnerable to nontargeted UAPs (Table 1). Specifically, the fooling rate, $R_f$, of the UAPs with $\zeta = 1\%$ for the test image set was 81.0% at most. A higher $\zeta$ led to a higher $R_f$. We observed that $R_f$ of the UAP with $\zeta = 2\%$ for the test image set was between 85.7% and 87.4%. Furthermore, the random UAPs with $\zeta = 2\%$ misclassified the models; specifically, their $R_f$ was 22.1% at most. The change in $R_f$ did not exhibit significant dependence on the norm types ($p = 2$ or $\infty$). The difference in $R_f$ for the test image set between $p = 2$ and $p = \infty$ was approximately 7% at most, when the model and the other parameters were the same. $R_f$ of the UAP against the COVIDNet-CXR Small model was lower than that of the COVIDNet-CXR Large model in the case of $\zeta = 1\%$, when the model and the other parameters were the same; however, no remarkable difference in $R_f$ between these models was observed in the case of $\zeta = 2\%$. $R_f$ for the training image set was higher than that for the test image set because the UAPs were generated based on the training image set.

**Table 1. Fooling rates $R_f$ (%) of nontargeted UAPs against the COVID-Net models.** $R_f$ for training and test images are presented. The values in the brackets indicate $R_f$ of random UAPs (random controls).

| $p$ | $\zeta$ | COVIDNet-CXR Small | | COVIDNet-CXR Large | |
|---|---|---|---|---|---|
| | | Training | Test | Training | Test |
| 2 | 1% | 61.4 (1.3) | 58.0 (0.4) | 90.0 (2.5) | 81.0 (3.9) |
| | 2% | 98.5 (12.6) | 87.4 (16.0) | 97.4 (17.9) | 85.7 (22.1) |
| ∞ | 1% | 70.8 (1.0) | 64.9 (1.3) | 84.8 (2.0) | 77.1 (3.5) |
| | 2% | 98.5 (9.4) | 87.4 (13.4) | 97.4 (14.3) | 85.7 (19.9) |

Due to nontargeted UAPs, the models classified most images into *COVID-19*. Figure 1 shows the confusion matrices for the COVID-Net models attacked using non-targeted UPAs with $p = \infty$. For the UAPs with $\zeta = 1\%$, the COVIDNet-CXR Small model classified >70% of the *normal* and *pneumonia* test images into *COVID-19*. Moreover, the COVIDNet-CXR Large model classified approximately 90% of the *normal* and *pneumonia* images into *COVID-19*. For a higher $\zeta$, this tendency was more significant. In particular, the COVIDNet-CXR Small and Large models judged almost all *normal* and *pneumonia* test images as COVID-19 cases when $\zeta = 2\%$. Additionally, the tendency of adversarial images being classified into *COVID-19* was observed when considering UAPs with $p = 2$ and the training image set.

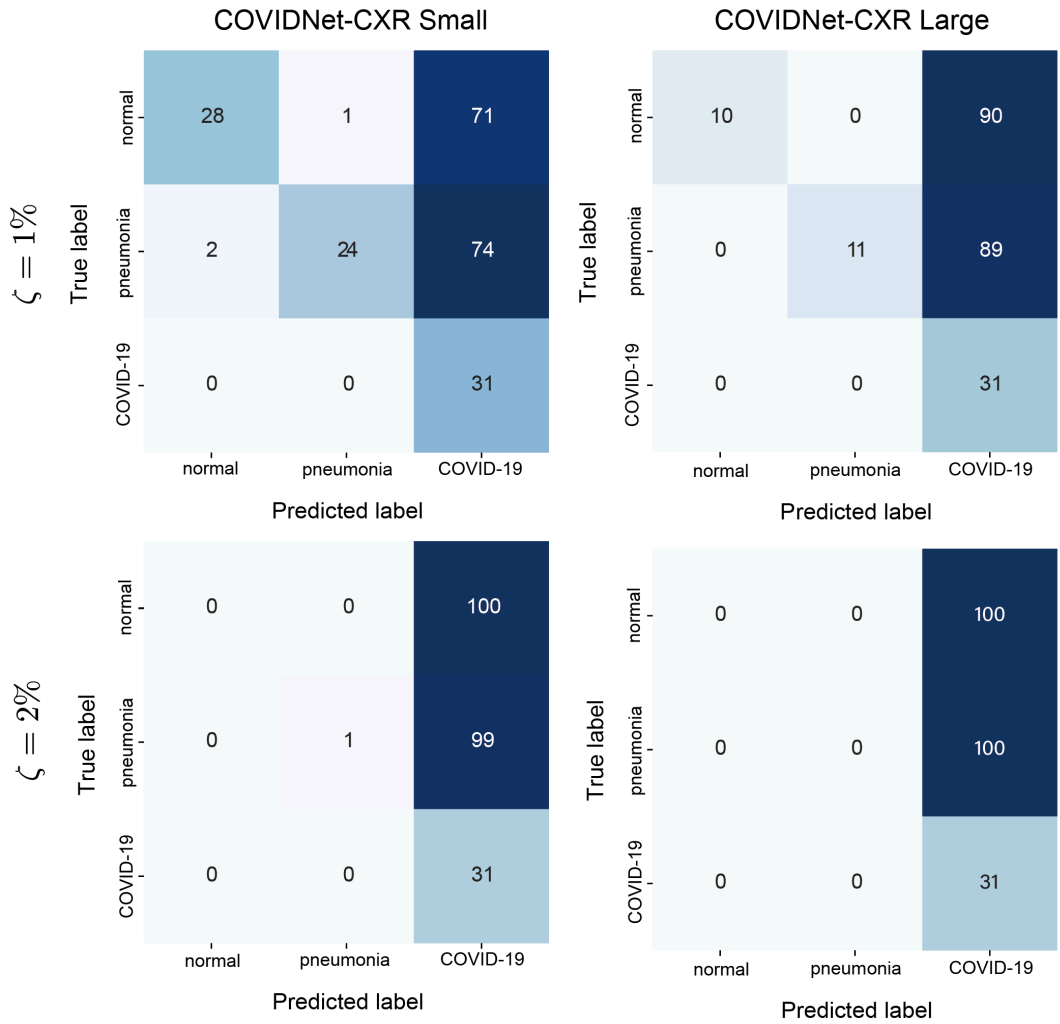

**Figure 1. Confusion matrices for the COVID-Net models attacked using the nontargeted UPAs on the test images.** $p = \infty$. Left and right panels represent the COVIDNet-CXR Small and COVIDNet-CXR Large models, respectively. The top and bottom panels indicate $\zeta = 1\%$ and $\zeta = 2\%$, respectively.

The nontargeted UAPs with $\zeta = 1\%$ and $\zeta = 2\%$ were almost imperceptible. Figure 2 shows the nontargeted UPAs with $p = \infty$ against the COVID-Net models and their adversarial images. The models classified the original X-ray images (left panels in Fig. 2) and correctly predicted their actual classes; however, they judged all adversarial images as COVID-19 cases due to the nontargeted UAPs. Similarly, the nontargeted UPAs with $p = 2$ were almost imperceptible.

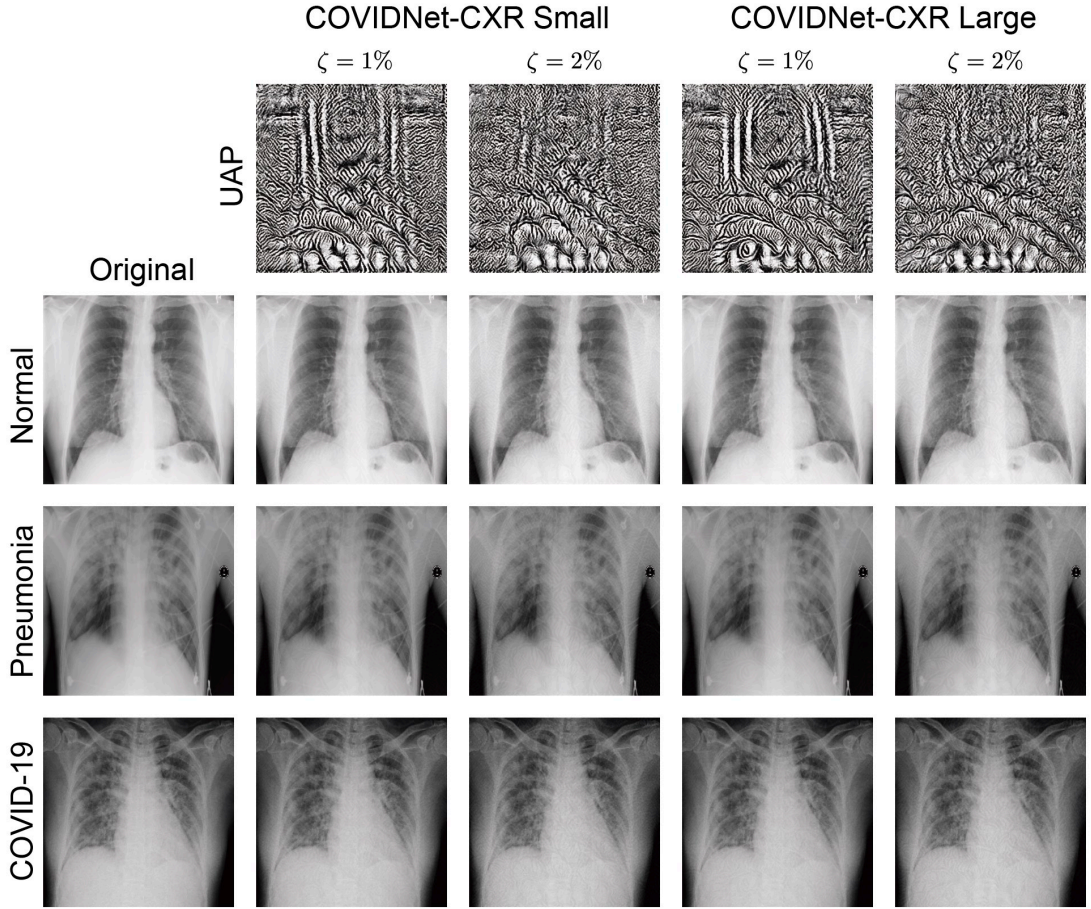

**Figure 2. Nontargeted UAPs with $p = \infty$ against the COVID-Net models and their adversarial images.** UAPs (top panels) with $\zeta = 1\%$ and $\zeta = 2\%$ are shown. The models correctly classified the original images (left panels) into their actual labels. The predicted labels of all adversarial images are *COVID-19*. Note that the UAPs are emphatically displayed for clarity; in particular, each UAP is scaled by a maximum of 1 and minimum of 0.

*3.3. Vulnerability to targeted universal adversarial perturbations*

Furthermore, we found that both COVIDNet-CXR Small model (Table 2) and COVIDNet-CXR Large model (Table 3) were vulnerable to targeted UAPs. Subsequently, we considered the effect of the targeted attacks using UAPs to each class: *normal*, *Pneumonia*, and *COVID-19*. When $\zeta = 1\%$, the targeted attack success rates, $R_s$, for the test images were between approximately 60% and 85% and between approximately 55% and 95% for the COVIDNet-CXR Small and Large models, respectively. Conversely, $R_s$ for the training images were between approximately 65% and 90% and between approximately 55% and 90%. $R_s$ of the UAP with $p = 2$ was generally higher than that of the UAP with $p = \infty$, when the model and the other parameters were the same. Moreover, no remarkable difference in $R_s$ was observed between the target classes; however, $R_s$ of the targeted attacks to *COVID-19* was relatively high in the COVIDNet-CXR Large model.

Thus, a higher $\zeta$ led to a higher $R_s$. When $\zeta = 2\%$, the $R_s$ values for both the training and test images were approximately 100%, regardless of the target classes. For the targeted attacks to *normal* and *pneumonia*, $R_s$ of random UAPs for the test images were also relatively high; in particular, they were between approximately 35% and 45% and between approximately 30% and 45% for the COVIDNet-CXR Small model and COVIDNet-CXR Large model, respectively.

**Table 2. Targeted attack success rate $R_s$ (%) of targeted UAPs against the COVIDNet-CXR Small model to each target class.** $R_s$ for training and test images are shown. The values in brackets are $R_s$ of random UAPs (random controls).

| $p$ | $\zeta$ | *Normal* | | *Pneumonia* | | *COVID-19* | |
|---|---|---|---|---|---|---|---|
| | | Training | Test | Training | Test | Training | Test |
| 2 | 1% | 88.1 (60.5) | 78.4 (46.3) | 76.7 (37.5) | 71.4 (41.6) | 68.1 (1.9) | 74.0 (12.1) |
| | 2% | 99.4 (54.4) | 97.8 (39.0) | 99.4 (33.0) | 98.7 (35.9) | 100 (12.6) | 99.1 (25.1) |
| $\infty$ | 1% | 79.5 (60.7) | 64.9 (45.9) | 66.5 (37.5) | 61.9 (41.6) | 78.8 (1.8) | 84.0 (12.6) |
| | 2% | 98.7 (56.3) | 96.1 (39.4) | 99.5 (34.1) | 98.3 (37.7) | 100 (9.5) | 100 (22.9) |

**Table 3. Targeted attack success rates $R_s$ (%) of targeted UAPs against the COVIDNet-CXR Large model to each target class.** $R_s$ for training and test images are shown. The values in brackets are $R_s$ of random UAPs (random controls).

| $p$ | $\zeta$ | *Normal* | | *Pneumonia* | | *COVID-19* | |
|---|---|---|---|---|---|---|---|
| | | Training | Test | Training | Test | Training | Test |
| 2 | 1% | 85.2 (58.9) | 71.4 (44.2) | 72.6 (37.0) | 66.2 (39.0) | 92.4 (4.0) | 95.2 (16.9) |
| | 2% | 99.2 (50.7) | 98.3 (34.6) | 99.5 (30.6) | 98.7 (32.9) | 100 (18.7) | 100 (32.5) |
| $\infty$ | 1% | 71.0 (59.2) | 56.7 (44.2) | 55.4 (37.0) | 53.2 (40.3) | 88.4 (3.7) | 92.2 (15.6) |
| | 2% | 97.9 (52.7) | 93.9 (35.9) | 99.4 (32.3) | 98.3 (33.8) | 100 (14.9) | 100 (30.3) |

It was difficult to classify the *COVID-19* images into another targeted class (*normal* or *pneumonia*) when UAPs were relatively weak (i.e., $\zeta = 1\%$). Figure 3 shows the confusion matrices for the COVIDNet-CXR Small model attacked using targeted UPAs with $p = \infty$. For both targeted attacks to *normal* and *pneumonia*, the model correctly predicted almost all *COVID-19* images as COVID-19 cases, despite the targeted attacks. Conversely, approximately 50% of *normal* (*pneumonia*) images were classified into the targeted class *pneumonia* (*normal*). However, for a higher $\zeta$ (i.e., $\zeta = 2\%$), the targeted attacks of the *COVID-19* images were successful; in particular, almost all *COVID-19* images were classified into the target class (*normal* or *pneumonia*) due to the UAP. The classification of the images into COVID-19 using targeted UAPs was easier than that into the other classes. Due to the UAP with $\zeta = 1\%$, the model judged approximately 80% of *normal* and *pneumonia* images as COVID-19 cases, respectively. Similar tendencies were observed in the COVIDNet-CXR Large model, for targeted UPAs with $p = 2$, and on the training image set.

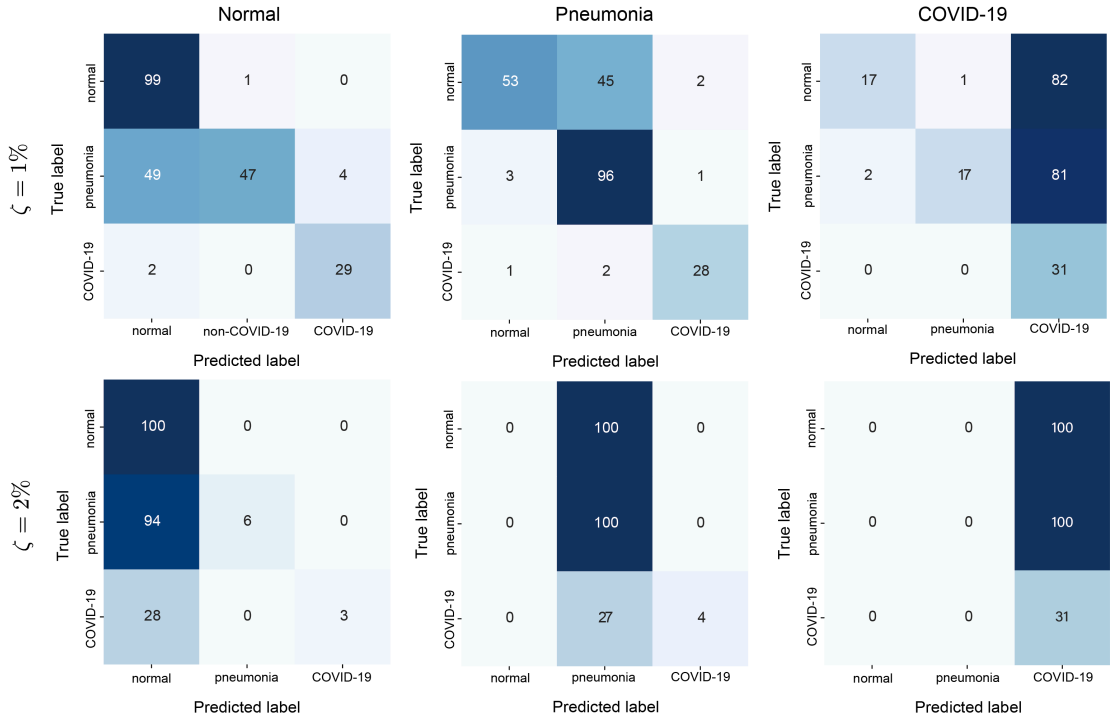

**Figure 3. Confusion matrices for the COVIDNet-CXR Small model attacked with the targeted UPAs with $p = \infty$ on the test images.** The left, middle, and right panels represent the targeted classes: *normal*, *pneumonia*, and *COVID-19*, respectively. The top and bottom panels indicate $\zeta = 1\%$ and $\zeta = 2\%$, respectively.

The targeted UAPs were also almost imperceptible. Figure 2 shows the targeted UPAs with $p = \infty$ and $\zeta = 2\%$ against the COVIDNet-CXR Small model and their adversarial images. The model classified the original images (left panels in Fig. 4) and correctly predicted their actual classes (source classes); however, it classified the adversarial images into each target class due to the targeted UAPs. The UAPs with $\zeta = 1\%$ were also imperceptible. Additionally, the imperceptibility was confirmed in the UAPs with $p = 2$ and those against the COVIDNet-CXR Large model.

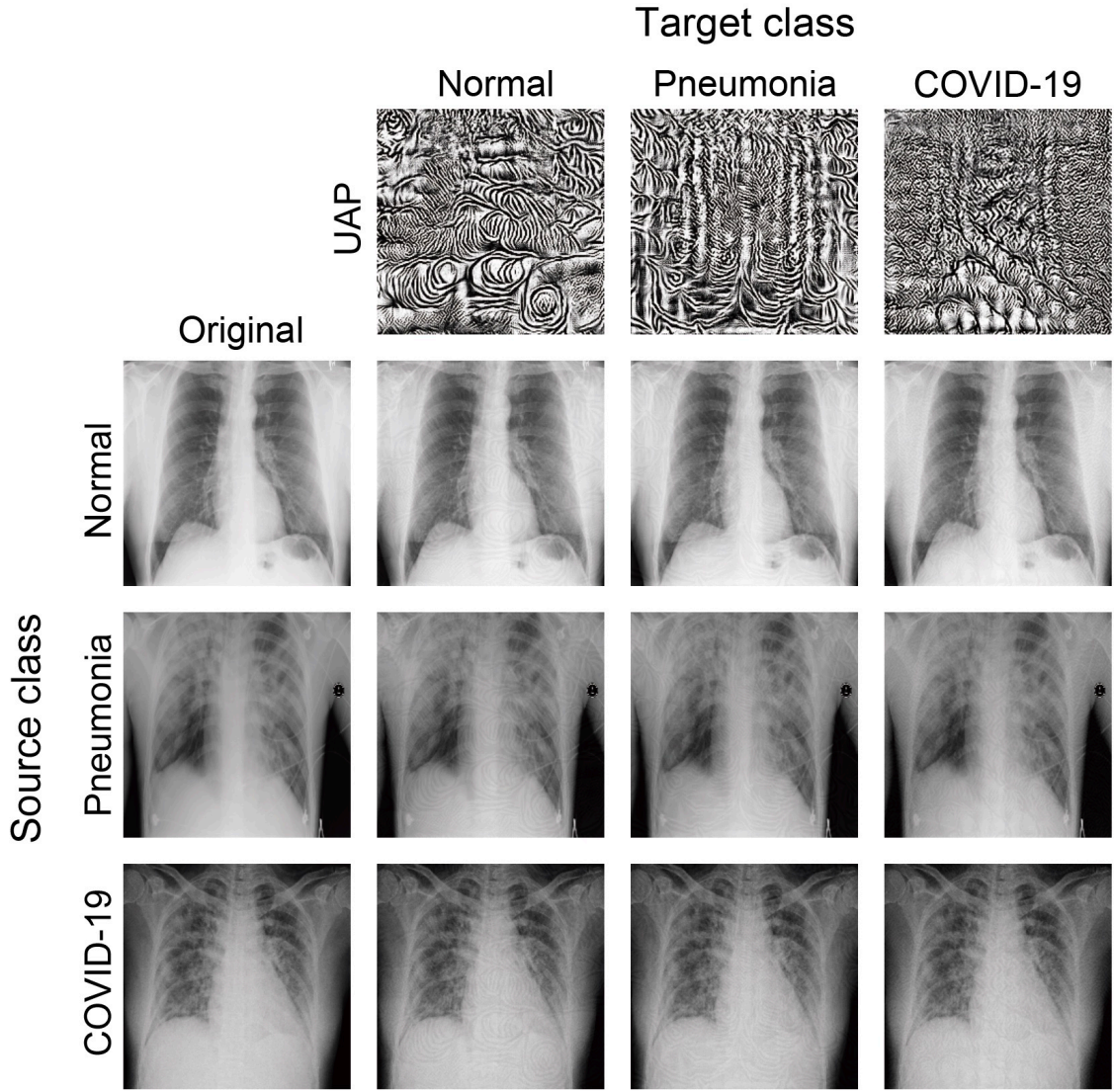

**Figure 4. Targeted UAPs (top panel) with $\zeta = 2\%$ and $p = \infty$ against the COVIDNet-CXR Small model and their adversarial images.** Note that the UAPs are emphatically displayed for clarity; in particular, each UAP is scaled by a maximum of 1 and minimum of 0.

*3.4. Effect of adversarial retraining*

Adversarial retraining is often used to avoid adversarial attacks. We investigated the extent to which adversarial retraining increases the robustness of the COVIDNet-CXR Small model to nontargeted and targeted UAPs with $p = \infty$. Adversarial retraining did not affect the test accuracy in both nontargeted and targeted cases; specifically, the accuracy on the (clean) test images remained constant at approximately 90% (Figs. 5A and 5B).

For nontargeted attacks using UAPs with $\zeta = 2\%$, $R_f$ for the test images declined with the iterations for adversarial retraining; in particular, it was 22.1% after five iterations (Fig.

5A). The confusion matrix (Fig. 5C) for the fine-tuned model obtained after five iterations indicates that the *normal* and *COVID-19* images were almost correctly classified despite the nontargeted UAPs, but 45% of the *pneumonia* images were still misclassified.

For targeted attacks to *COVID-19* using UAPs with $\zeta = 1\%$, $R_s$ for the test images decreased with the iterations for adversarial retraining (Fig. 5B); specifically, it was 16.5% after five iterations. The confusion matrix (Fig. 5D) for the fine-tuned model obtained after five iterations indicates that the *normal* and *COVID-19* images were almost correctly classified despite the targeted UAPs, but 15% of the *pneumonia* images were still misclassified into *COVID-19*.

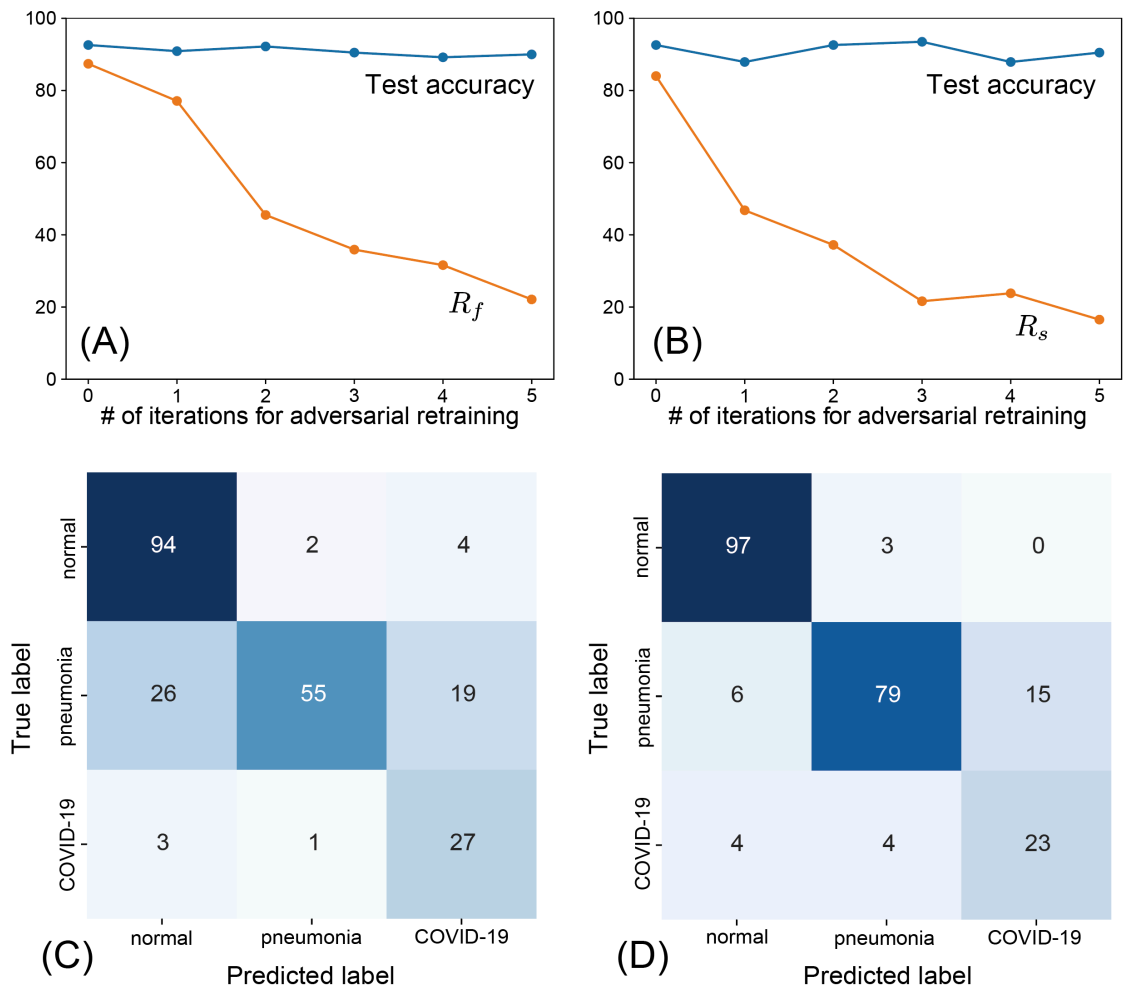

**Figure 5. Effect of adversarial retraining on the robustness to UAPs with $p = \infty$ against the COVIDNet-CXR Small model.** Scatter plots of (A) the fooling rate, $R_f$ (%), for nontargeted UAPs with $\zeta = 2\%$ versus the number, $N_i$, of iterations for adversarial retraining and (B) the targeted attack success rate, $R_s$ (%), of targeted UAPs with $\zeta = 1\%$ to *COVID-19* versus $N_i$. $R_f$ and $R_s$ are for the test images. The accuracies (%) on the set of clean test images are also shown. The confusion matrices for the fine-tuned models obtained after five iterations of adversarial retraining using the (C) nontargeted

UAPs and (D) targeted UAPs. Note that these confusion matrices are for the fine-tuned models attacked using nontargeted and targeted UAPs, respectively.

## 4. Discussion

The COVID-Net models were vulnerable to small UAPs; moreover, they were slightly less robust to random UAPs. The results indicated that the DNN-based systems were easy to fool. Given that the vulnerability to UAPs is observed in various DNN architectures [23,24], they are expected to universally exist in DNN-based systems for detecting COVID-19 cases.

For nontargeted attacks, the COVID-Net models predicted most of the chest X-ray images as COVID-19 cases due the UAPs (Fig. 1), although the UAPs were almost imperceptible (Fig. 2). This result is consistent with the tendency that DNN models classify most inputs into a few specific classes due to nontargeted UAPs (i.e., existence of dominant labels in nontargeted attacks based on a UAP) [23]. moreover, this indicates that the models provide false positives in COVID-19 diagnosis, which may cause unwanted mental stress to patients and complicate the estimation of the number of COVID-19 cases. The dominant label of COVID-19, observed in this study, may be because the COVIDx dataset was imbalanced. The images in *COVID-19* are dominantly fewer than those in *normal* and *pneumonia*. The algorithm considers maximizing the fooling rate; thus, a relatively large fooling rate is achieved when all inputs are classified into *COVID-19* due to UAPs. In addition, the observed dominant label may be because the losses were computed by weighting to *COVID-19* class. The decision for the *COVID-19* class might be more susceptive to changes in pixel values than that for the other classes.

The relatively easy targeted attacks to *COVID-19* (Fig. 3) may be because *COVID-19* was the dominant label. Moreover, the targeted attacks to *normal* and *pneumonia* were possible, despite almost imperceptible UAPs (Fig. 4). The results imply that a third party can control the DNN-based systems, which may lead to security concerns. The targeted attacks cause both false positives and negatives, and thus, can be used to adjust the number of COVID-19 cases. Moreover, they may affect individual and social awareness of COVID-19 (e.g., voluntary restraint and social distancing). These may lead to problems in terms of public health (i.e., minimizing the spread of the pandemic) and economy. More generally, complex classifiers, including DNNs, are currently used for high-stake decision-making in healthcare; however, they can potentially cause catastrophic harm to society because they are often difficult to interpret [30].

A simple solution to avoid these problems is to make DNN-based systems closed-source and publicly unavailable; however, this conflicts with the purpose of accelerating the development of computer-based systems for detecting COVID-19 cases and COVID-19 treatment. An alternative may be to consider black-box systems, such as closed application programming interfaces (APIs) and closed-source software in which only queries on inputs are allowed and outputs are accessible. Such closed APIs are better because they are at least publicly available. However, it is possible that the APIs are vulnerable to adversarial attacks. This is because UAPs have generalizability [23] (i.e., UAPs for a DNN can fool another DNN). That is, the adversarial attacks on black-box DNN-based systems may be possible using UAPs generated based on white-box DNNs. Moreover, several methods for adversarial attacks on black-box DNN-based systems, which estimate adversarial perturbations using only model outputs (e.g., confidence scores), have been proposed [31–33].

Therefore, defense strategies against adversarial attacks should be considered. A simple defense strategy is to fine-tune DNN models using adversarial images [22,23,27]. In fact, we demonstrated that iterative fine-tuning of a DNN model using UAPs improved the robustness of the DNN model to nontargeted and targeted UAPs (Fig. 5). However, the iterative fine-tuning method required high computational costs; moreover, it did not perfectly avoid vulnerability to UAPs. In addition, several methods breaching defenses using adversarial retraining have already been proposed [27]. Alternatively, dimensionality reduction (e.g., principle component analysis), distributional detection (e.g., maximum mean discrepancy), and normalization detection (e.g., dropout randomization) may be useful for adversarial defenses; however, adversarial examples are not easily detected using these approaches [27]. Defending against adversarial attacks is a cat-and-mouse game [26]; thus, it may be difficult to completely avoid security concerns caused by adversarial attacks. However, the development of methods for defending against adversarial attacks has been advanced. For example, detecting adversarial attack-based robustness to random noise [34], use of a discontinuous activation function that purposely invalidates the DNN's gradient at densely distributed input data points [35], and DNNs for purifying adversarial examples [36] may help reduce the concerns.

In conclusion, we demonstrated the vulnerability of DNNs for detecting COVID-19 cases to nontargeted and targeted attacks based on UAPs. Our findings emphasize that careful consideration is required in developing DNN-based systems for detecting COVID-19 cases and their practical applications.

**Ethics.** This study required no ethical permit.

**Data availability.** The code used in this study is available from our GitHub repository: github.com/hkthirano/UAP-COVID-Net. The chest X-ray images used in this study are publicly available online (see github.com/lindawangg/COVID-Net/blob/master/docs/COVIDx.md for details).

**Authors' contributions.** KT conceived and designed the study. HH and KK prepared the data and models. HH coded and performed experimental evaluation. HH and KT interpreted the results. HH and KT wrote the manuscript. All authors gave the final approval for publication.

**Competing interests.** The authors have declared no competing interest.

**Funding.** No specific funding was awarded for this research.

**Acknowledgements.** The authors are much obliged to Dr. Seyed-Mohsen Moosavi-Dezfooli for his helpful comments regarding fine-tuning of DNN models with UAPs. The authors would like to thank Editage (www.editage.com) for English language editing.